\pdfoutput=1

\documentclass[11pt]{article}

\usepackage[final]{acl}

\usepackage{times}
\usepackage{latexsym}

\usepackage[T1]{fontenc}

\usepackage[utf8]{inputenc}

\usepackage{microtype}

\usepackage{inconsolata}

\usepackage{graphicx}

\usepackage{booktabs}
\usepackage{multirow}

\title{Selecting and Merging: Towards Adaptable and Scalable Named Entity Recognition with Large Language Models}

\author{
Zhuojun Ding\textsuperscript{1}, 
Wei Wei\thanks{Corresponding author}\textsuperscript{1}
\and Chenghao Fan\textsuperscript{1} \\
\textsuperscript{1} School of Computer Science \& Technology, Huazhong University of Science and Technology \\
\texttt{\{dingzj, weiw\}@hust.edu.cn, facicofan@gmail.com}
}

\begin{document}

\maketitle

\begin{abstract}

Supervised fine-tuning (SFT) is widely used to align large language models (LLMs) with information extraction (IE) tasks, such as named entity recognition (NER). 
However, annotating such fine-grained labels and training domain-specific models is costly. 
Existing works typically train a unified model across multiple domains, but such approaches lack adaptation and scalability since not all training data benefits target domains and scaling trained models remains challenging.
We propose the SaM framework, which dynamically \textbf{S}elects \textbf{a}nd \textbf{M}erges expert models at inference time.
Specifically, for a target domain, we select domain-specific experts pre-trained on existing domains based on (i) domain similarity to the target domain and (ii) performance on sampled instances, respectively. The experts are then merged to create task-specific models optimized for the target domain.
By dynamically merging experts beneficial to target domains, we improve generalization across various domains without extra training. Additionally, experts can be added or removed conveniently, leading to great scalability.
Extensive experiments on multiple benchmarks demonstrate our framework’s effectiveness, which outperforms the unified model by an average of 10\%. 
We further provide insights into potential improvements, practical experience, and extensions of our framework.\footnote{\url{https://github.com/Ding-ZJ/SaM}}

\end{abstract}

\section{Introduction}

\begin{figure}[t]
  \centering
  \includegraphics[width=1.0\linewidth]{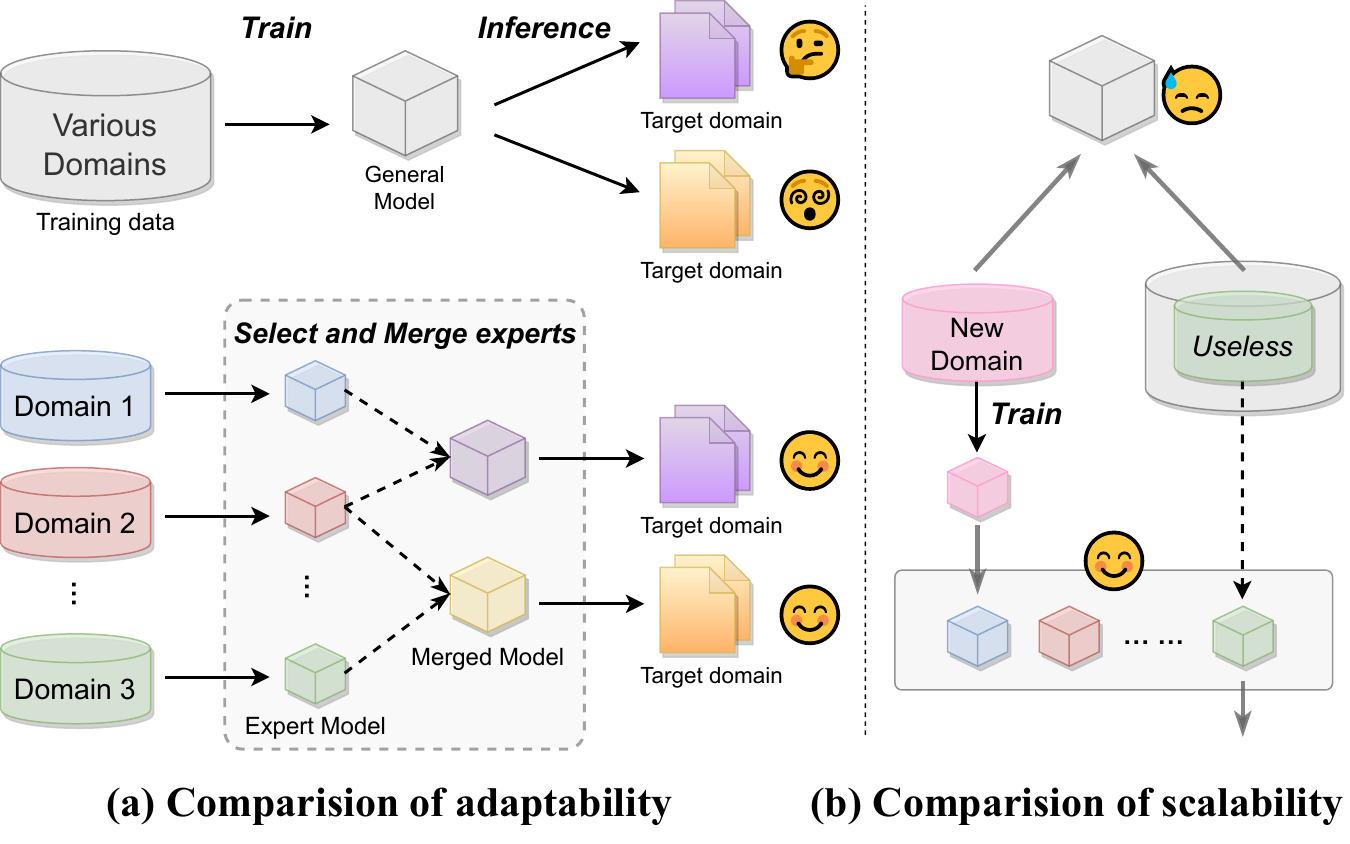}
  \caption{(a) Existing methods train a general unified model across multiple domains, while we dynamically select and merge expert models at inference time. (b) A trained system struggles to accommodate changes in training data, while we flexibly add or remove expert models, ensuring great scalability.}
  \label{fig:intro}
  \vspace{-4mm}
\end{figure}

Large language models (LLMs) demonstrate remarkable performance across a wide range of tasks~\citep{achiam2023gpt,yang2024qwen2,guo2025deepseek}, but still struggle with information extraction (IE) tasks~\citep{xu2024large,ding2024improving,Fan_Wei_Qu_Lu_Xie_Cheng_Chen_2024}, such as Named Entity Recognition (NER). The inherent gap between task formulations and LLM training objectives is a critical factor underlying this limitation.
To mitigate this, supervised fine-tuning (SFT) has become a widely used strategy, demonstrating significant improvements~\citep{wang2023instructuie,zhou2024universalner,fan2025makeloragreatagain}.

However, annotating data and training domain-specific models each time is costly, particularly for fine-grained IE tasks. Most existing approaches collect large-scale training data from multiple domains to train a unified model~\citep{wang2023instructuie,sainz2024gollie,yang2025b2ner}. 
Although such models exhibit cross-domain generalization capabilities, they frequently exhibit suboptimal performance in both in-domain and out-of-domain test scenarios. This limitation arises primarily because \textcolor{black}{(1) not all training samples universally enhance performance on a given target domain~\citep{liu2024makes,zhou2024lima}, and (2) inherent conflicts may emerge across heterogeneous domains during joint training, leading to compromised optimization efficacy~\citep{sainz2024gollie,yang2025b2ner,fan2024on}.}
Additionally, even when data from the target domain is available, effectively integrating it into a trained model without compromising performance remains challenging.

To address these issues, we adopt a model merging strategy~\citep{il2023taskArith} to dynamically select domain-specific models for different target domains and fuse their parameters to obtain task-specific models. 
Specifically, we first train multiple expert models in different domains with available data. 
Then, we design the \textbf{SaM} framework to derive task-specific models by \textbf{S}electing \textbf{a}nd \textbf{M}erging experts from two perspectives: (i) Domain similarity. We assess the domain similarity between the target domain and each expert model, and select the most relevant experts for parameter fusion to create a task-specific model tailored to the target domain. (ii) Sampling evaluation. We randomly sample \textit{k} data instances from the target domain and integrate the predictions of all experts as pseudo-labels (no ground truth labels required). Based on these labels, we assess the performance of each expert on the sampled data and select the best-performing experts to merge into another task-specific model.
The two perspectives synergistically complement each other: domain similarity provides a high-level, coarse-grained assessment that establishes theoretical priors, while sampling evaluation delivers a fine-grained, empirical quantification of expert performance to yield actionable practical guidance.
By integrating the outputs of both task-specific models, we achieve more comprehensive results.

Compared to previous methods (Figure \ref{fig:intro}), we allow for task-specific customization across diverse target domains, providing improved generalization ability without extra training. It also provides great scalability, as experts can be easily added or removed based on practical needs.
Notably, our approach is orthogonal to previous studies. By leveraging their practical insights, we can train more effective expert models, thereby enhancing the performance of our framework.
In terms of resource requirements, our approach does not incur additional training costs. By leveraging parameter-efficient fine-tuning methods~\cite{hu2022lora}, we only introduce minimal storage overhead. 
\textcolor{black}{Additionally, by employing either strategy individually or further integrating the target models derived from both strategies, we can achieve comparable performance without incurring additional inference costs.}

In summary, our contributions are as follows: \textbf{(1)} We introduce a model-merging paradigm for LLM-based Named Entity Recognition, enhancing adaptability and scalability. \textbf{(2)} We propose a model selection strategy based on domain similarity and sampling evaluation, which effectively selects expert models beneficial to the target domain for merging. \textbf{(3)} Experimental results demonstrate the effectiveness of our framework, which outperforms the unified model by an average of 10\% and by up to 20\% in certain domains.
Further experiments analyze potential improvements, practical experience, and framework generalizability, providing deeper practical insights.

\section{Related Works}

\paragraph{LLMs for Information Extraction}

Current LLMs-based IE mainly fall into two paradigms.

One paradigm uses larger models. Training them needs significant computational resources, and fine-tuning them specifically for IE tasks may be not cost-effective. However, these models excel in instruction-following and reasoning. Therefore, such methods focus on optimizing task instructions, reasoning strategies, or in-context learning (ICL) demonstrations. \citet{li2023codeie} show that code-style prompts enhance IE tasks. \citet{pang2023guideline} and \citet{tong2025evoprompt} prompt LLMs with more comprehensive information to improve task understanding. \citet{xie2023empirical} and \citet{wan2023gptRE} introduce reasoning techniques such as Chain-of-Thought (CoT) to guide the model in step-by-step task completion. \citet{xie2024self} employ self-consistency to generate reliable ICL examples.

Another paradigm uses smaller models. While these models have weaker instruction-following capabilities, they require much fewer training resources. Such methods enhance LLMs through supervised fine-tuning~\citep{wang2023instructuie}.
Many studies design optimization strategies on the data side. \citet{yang2025b2ner} resolves conflicts and redundancy in training data. \citet{zhou2024universalner} distills more diverse data from ChatGPT. \citet{li2024knowcoder} formats training data in code style. \citet{sainz2024gollie} enriches instructions with detailed task descriptions. \citet{ding2024rethinking} emphasizes negative samples. In addition to instruction tuning, \citet{qi2024adelie} further employs alignment training~\citep{rafailov2024dpo}, and \citet{guo2025baner} incorporates contrastive learning objectives.
Instead of training one universal model, we train several domain experts and design a merging method to improve adaptability and scalability.

Additionally, the backbone model is also critical. Recently, code-based LLMs have gained popularity, as they may better suit IE tasks than natural language-based LLMs~\citep{li2023codeie}.

\paragraph{Model Merging}

Model merging integrates multiple task-specific models at the parameter level to create a unified model, which could handle multiple tasks simultaneously and exhibit better out-of-domain generalization. Unlike multi-task learning, model merging reuses existing models, reducing computational and data demands since only model parameters are needed.
Beyond simple parameter averaging, \citet{matena2022fisherMerging} assigns different importance to model parameters. \citet{il2023taskArith} applies arithmetic operations for finer control over model behavior. \citet{jin2023regmean} enforces output consistency between the merged model and its constituent models. \citet{yu2024dare} and \citet{yadav2023ties} mitigate inter-model interference by addressing parameter redundancy or sign inconsistency, and weight sparsity, respectively. \citet{lu2024twin} decomposes model parameters into shared and task-specific components.
In this paper, we introduce model merging to improve adaptability and scalability across different target domains.

\begin{figure*}[th]
  \centering
  \includegraphics[width=0.99\linewidth]{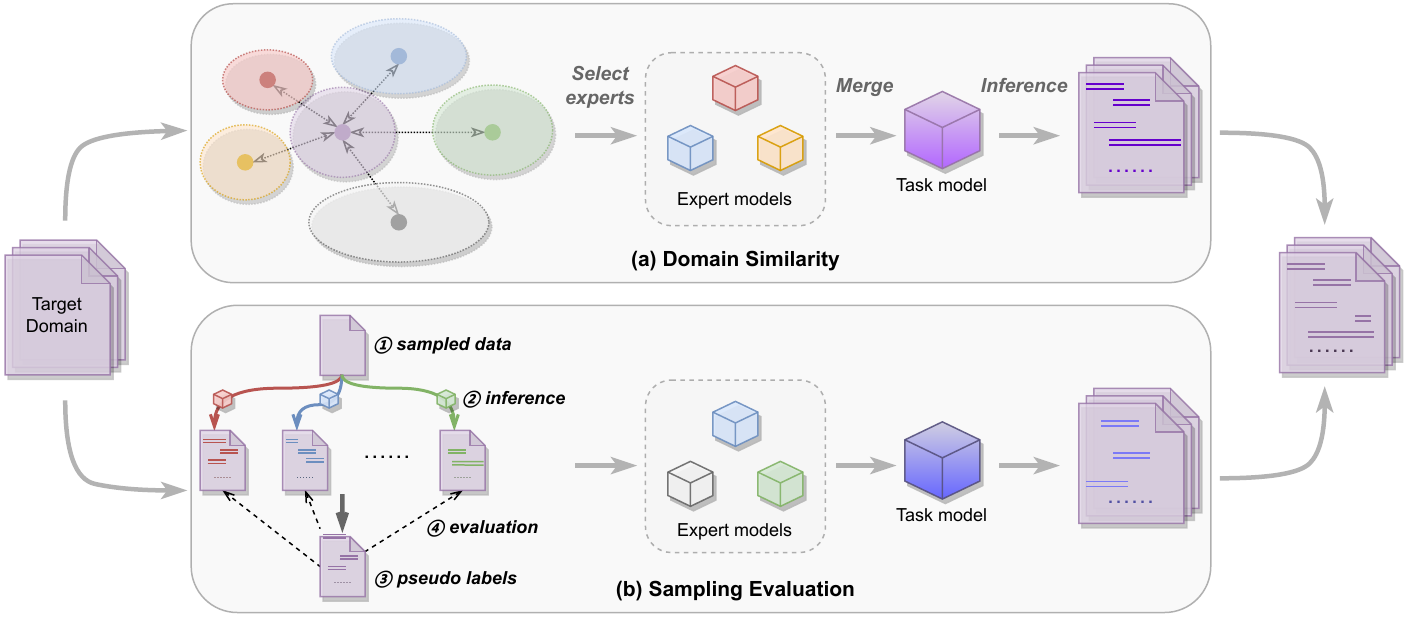}
  \caption{Framework overview. Given a target domain, we select expert models from two perspectives: 
  \textbf{(a) Domain Similarity}, which selects experts from domains most similar to the target domain. We compute the centroid of all data embeddings as the domain embedding and measure similarities by cosine distance. 
  \textbf{(b) Sampling Evaluation}, which selects experts with better performances on sampled instances from the target domain. To reduce reliance on ground-truth labels, we ensemble predictions from all experts as (pseudo) labels.
  We merge models within each expert subset to obtain two task-specific models.
  The final result integrates the outputs of the two task models.}
  \label{fig:framework}
\end{figure*}

\section{Methodology}

In this section, we first provide an overview of the training process of domain expert models. Then we explain our SaM framework for selecting and merging experts, as shown in Figure~\ref{fig:framework}.

\subsection{Training Domain Experts}
\paragraph{Data Collection.}
\label{method_data_collection}

We first collect more than 20 commonly used NER datasets and classify them into six domains based on their sources: News, Social media, Biomedical, STEM (Science, Technology, Engineering, and Mathematics), Legal, and Traffic. 
We remove 90\% of the NA data that contains no entities. Through sampling or redundancy, we limit the total samples per domain to between 10,000 and 50,000. The number of sampled instances from each dataset was proportional to the number of entity types it contained.
Detailed information is provided in Appendix~\ref{apd:training_details}. 

\paragraph{Training Data Construction.}

Refering to practice of prior studies~\cite{wang2023instructuie,qi2024adelie}, we format the raw data into task instructions, inputs, and outputs for training. 

The task instructions consist of: \textbf{(1)} Data source description: A brief overview of the dataset source. \textbf{(2)} Entity type description: Concise definition of entity types. \textbf{(3)} In-context learning demonstrations: $1\sim5$ randomly selected input-output pairs from the training data. \textbf{(4)} Label drop: Excluding the requirement for recognizing certain entity types. \textbf{(5)} Label masking: Replacing entity labels with abstract placeholders such as ``Type1''. Empirically, we apply \textit{(1)}, \textit{(2)}, and \textit{(3)} to 70\% of the data; \textit{(4)} to 30\%; and \textit{(5)} to 5\%. These modifications are applied independently, except \textit{(5)}, which should co-occur with \textit{(2)}. For output parts, we adopt three formats: JSON (e.g., ``\{entity span: entity type\}''), enumeration (e.g., ``Type: span1, span2, ...''), and natural language descriptions.

These strategies help enhance model robustness to some extent. However, we claim this is not an optimal configuration, as our focus is not on training a best-performing model.

\paragraph{Model Training.}

Following prior work, we train the model using instruction tuning. Given a dataset $D_A = \{(I, X, Y)\}$ from domain $A$, where $I$ is the task instruction, $X$ is the input sequence, and $Y=\{y_i\}_{i=1}^L$ is the output sequence (i.e., entity predictions), the training loss of the domain-specific expert model $\mathcal{M}_A$ is defined as:
\begin{equation}
    \mathcal{L}_{\theta_A} = -\sum_{D_A}\sum_{t=1}^{L}\log P_{\theta_A}(y_t\mid I,X,y_{<t})
\end{equation}
where $\theta_A$ denotes the parameters of $\mathcal{M}_A$.

\subsection{Selecting and Merging Experts}
\label{method:SaM}

When handling a specific target domain, we select a subset of expert models and fuse their parameters to obtain task-specific models. As shown in Figure~\ref{fig:framework}, this process is conducted from two perspectives.

\paragraph{Selecting with Domain Similarity.}

Given a domain with raw data $D_A = \{x_i\}$, we obtain the corresponding data embeddings $H_A=\{h_i\}$ through a text encoder. The domain embedding is then defined as the centroid of these data embeddings:  
\begin{equation}
    h_A = \frac{1}{|H_A|} \sum_{h_i \in H_A} h_i
\end{equation}

We compute domain embeddings $\{h_{e_i}\}$ for domains of all expert models and $h_t$ for the target domain. 
Then we compute the similarity between the expert domains and the target domain with cosine distances. Finally, the top-$m$ similar expert models are selected for model merging.  

The domain embedding inherently captures the data distribution in the embedding space. 
Thus, in theory, the selected expert models exhibit a certain degree of similarity and are expected to perform well in the target domain due to the resemblance in these data distributions.

\paragraph{Selecting with Sampling Evaluation.}

The selection of experts based on domain similarity is theoretically sound. However, in practice, we observe that the model with the highest domain similarity to the target domain does not always yield the best performance. Thus, we propose another selection strategy driven by model performance.

Specifically, we first randomly sample $k$ data instances from the target domain. Then each expert model generates predictions for them. To reduce dependence on ground-truth labels, we aggregate predictions via majority voting to construct pseudo-labels, which are subsequently used to assess expert performance. The top $m$ experts with the highest performance are selected for model merging.

Unlike domain similarity-based selection, this approach prioritizes practical effectiveness. As a result, the selected experts generally perform better individually in the target domain. However, we aim to obtain a superior task model through model merging, where individual performance is not the sole determining factor.

\paragraph{Merging Experts.}

Given a base model $\mathcal{M}_{base}$ and the supervised fine-tuned model $\mathcal{M}_{sft}$, we denote their parameters as $\theta_{base}$ and $\theta_{sft}$, respectively. The delta parameter $\delta_{sft}=\theta_{sft}-\theta_{base}$ serves as a parametric representation of the model's learned capabilities and is also referred to as the task vector.
Given multiple task-specific models $\{\mathcal{M}_{sft_i}\}$, we can merge them into a unified model $\mathcal{M}_{merge}$ with diverse capabilities~\cite{matena2022fisherMerging, il2023taskArith}:
\begin{equation}
    \theta_{merge} = \theta_{base} + \mathrm{Merge}(\delta_{sft_1},\delta_{sft_2},\cdots)
\end{equation}
where $\mathrm{Merge}(\cdot)$ denotes the model merging technique, such as simple averaging and task arithmetic~\cite{il2023taskArith}. We employ the Ties-Merging~\cite{yadav2023ties} method, which addresses parameter redundancy and sign inconsistency to mitigate inter-model interference when merging multiple models.
We selected two sets of expert models based on Domain Similarity (\textbf{DS}) and Sampling Evaluation (\textbf{SE}), respectively. These experts are subsequently merged to obtain two task-specific models, $\mathcal{M}_{DS}$ and $\mathcal{M}_{SE}$.

\begin{table*}[t!]
\centering
\resizebox{\textwidth}{!}{
\begin{tabular}{clcccccccc}
\toprule
 & & \multicolumn{5}{c}{\textbf{CrossNER}} & \multicolumn{2}{c}{\textbf{MIT}} & \multirow{2}{*}{\textbf{Average}} \\ 
\cmidrule(lr){3-7} \cmidrule(lr){8-9}
 & & AI & Literature & Music & Politics & Science & Movie & Restaurant & \\
\midrule
\multirow{6}{*}{\begin{tabular}[c]{c}\textbf{Recent}\\ \textbf{Studies}\end{tabular}} & InstructUIE & 48.40 & 48.80 & 54.40 & 49.90 & 49.40 & 63.00 & 20.99 & 47.84 \\
 & UniNER    & \textbf{62.90} & \underline{64.90} & 70.60 & 66.90 & \underline{70.80} & 61.20 & 35.20 & 61.79 \\
 & GoLLIE    & 59.10 & 62.70 & 67.80 & 57.20 & 55.50 & 63.00 & 43.40 & 58.39 \\
 & KnowCoder & 60.30 & 61.10 & 70.00 & 72.20 & 59.10 & 50.00 & 48.20 & 60.13 \\
& GLiNER    & 57.20 & 64.40 & 69.60 & \underline{72.60} & 62.60 & 57.20 & 42.90 & 60.90 \\
 & B2NER     & 59.00 & 63.70 & 68.60 & 67.80 & \textbf{72.00} & 67.60 & \textbf{53.30} & \underline{64.57} \\
\midrule
\multirow{2}{*}{\textbf{Fully-trained}} & Llama & 54.11 & 63.44 & \underline{72.53} & 62.92 & 59.09 & 64.81 & 49.68 & 60.94\\
 & Qwen & 51.38 & 53.46 & 61.12 & 54.99 & 59.39 & 66.66 & 52.69 & 57.10 \\
\midrule
\multirow{2}{*}{\textbf{SaM (Ours)}} & Llama & \underline{60.98}$_{12.7\%}$ & \textbf{66.93}$_{5.50\%}$ & \textbf{73.53}$_{1.4\%}$ & \textbf{74.47}$_{18.4\%}$ & 62.60$_{5.9\%}$ & \textbf{72.17}$_{11.4\%}$ & \underline{52.99}$_{6.7\%}$ & \textbf{66.24}$_{8.70\%}$ \\
 & Qwen & 60.01$_{15.8\%}$ & 61.99$_{16.0\%}$ & 65.93$_{7.9\%}$ & 67.05$_{21.9\%}$ & 62.41$_{5.1\%}$ & \underline{71.65}$_{7.50\%}$ & 52.90$_{0.4\%}$ & 63.13$_{10.6\%}$ \\
\bottomrule
\end{tabular}
}
\caption{Experimental results. We compare our method with recent studies and our fully-trained model 
(e.g., a unified model trained on all data we used). 
The best results are highlighted in bold, while suboptimal results are underlined. The right subscript denotes the percentage improvement compared to the fully trained model.}
\label{tab:main_results}
\end{table*}
\subsection{Inference}

For a target domain, we obtain two task-specific models, $\mathcal{M}_{DS}$ and $\mathcal{M}_{SE}$, following the methodology described in Section~\ref{method:SaM}. Each model independently generates predictions, producing two output sets, $Y_{DS}$ and $Y_{SE}$. 
Taking the intersection of two sets of predictions typically enhances reliability. However, our two task-specific models are already tailored for the target domain and capture different and complementary perspectives. Therefore, we adopt their union as the final result.

\section{Experiments}

\subsection{Setup}
\paragraph{Benchmarks, Baselines, and Metrics}
We evaluate our framework on two widely used benchmarks CrossNER~\citep{liu2021crossner} and MIT~\citep{ushio2021mit}, which contain datasets from seven domains (AI, Literature, Music, Politics, Science, Movie, and Restaurant) in total.
Our experiments are under zero-shot settings (i.e., no labeled target domain data), follow prior work. The source data for training and the target data for evaluation have different distributions.

We introduce two types of baselines for comparison. 
The first is the \textbf{Fully-trained} model, a single unified model trained on data from all domains using the same training configuration as us. This serves as the primary baseline to assess the effectiveness of our framework. 
The second includes recent studies that also train unified models but incorporate other advanced training optimizations, including \textbf{InstructUIE}~\citep{wang2023instructuie}, \textbf{UniNER}~\citep{zhou2024universalner}, \textbf{GoLLIE}~\citep{sainz2024gollie}, \textbf{KnowCoder}~\citep{li2024knowcoder}, 
\textbf{GLiNER}~\citep{zaratiana2024gliner},
and \textbf{B2NER}~\citep{yang2025b2ner}.
Most of these models use LLMs as the foundation, except for GLiNER, which contains only 300 million parameters.

Following prior studies~\cite{wang2023instructuie}, we use the entity-level micro-F1 score as the evaluation metric, where both the entity boundary and entity type should be correctly predicted.

\paragraph{Implementations}
We employ models from the Qwen and Llama series as base models for 
 our experiments. Specifically, we adopt the base version of  
Qwen2.5-7B and Llama3.1-8B as foundations and train expert models using LoRA~\citep{hu2022lora}. 
We employ the all-MiniLM-L6-v2\footnote{\url{https://huggingface.co/sentence-transformers/all-MiniLM-L6-v2}} text encoder to produce text embeddings.
We set $m$ (the number of selected models for merging) to 3.
We set $k$ (the number of sampled data instances) to 10.
More details are reported in Appendix~\ref{apd:training_details}.

\subsection{\textcolor{black}{Main Results}}

As shown in Table~\ref{tab:main_results}, our approach significantly outperforms the fully trained model across all target domains, achieving an average improvement of approximately 10\%, with gains of up to 20\% in specific domains. This demonstrates the effectiveness and superior domain adaptability of our approach.
To ensure practical comparability, we also compare our results with recent studies. 
These methods employ various training optimization strategies.
For example, B2NER mitigates redundant and conflicting information in the training data.
These techniques are orthogonal to ours. Notably, comparing these methods with our fully-trained model suggests that refining our training configuration could enhance our expert models and further improve the performance of our approach.
Our approach may incur slight computational and storage overhead, which is acceptable, as discussed in Appendix~\ref{appendix:cost}.

\begin{table*}[t]
\centering
\resizebox{0.97\textwidth}{!}{
\begin{tabular}{clcccccccc}
\toprule
 & & \textbf{AI} & \textbf{Literature} & \textbf{Music} & \textbf{Politics} & \textbf{Science} & \textbf{Movie} & \textbf{Restaurant} & \textbf{Average} \\
\midrule
\multirow{5}{*}{Llama} & w/o Merging & 53.72 & 59.95 & 71.61 & 71.88 & 59.09 & 66.43 & 50.62 & 61.90 \\
 & w/o Domain Similarity & 58.43 & 59.87 & 72.89 & \underline{74.20} & 59.92 & 68.91 & 51.69 & 63.70 \\
 & w/o Sampling Evaluation & 56.42 & \underline{63.92} & 72.25 & 72.99 & 61.71 & \underline{71.08} & \underline{52.48} & \underline{64.41} \\
 & w/o Selection & \underline{60.14} & 63.74 & \textbf{74.83} & 73.00 & \textbf{63.68} & 67.30 & 47.64 & 64.33 \\
\cmidrule(lr){2-10}
 & \textbf{SaM (Ours)} & \textbf{60.98} & \textbf{66.93} & \underline{73.53} & \textbf{74.47} & \underline{62.60} & \textbf{72.17} & \textbf{52.99} & \textbf{66.24} \\
\midrule
\multirow{5}{*}{Qwen} & w/o Merging & 57.90 & 59.01 & 65.33 & 65.73 & 60.21 & 65.45 & 51.11 & 60.68 \\
 & w/o Domain Similarity & \underline{58.57} & \underline{61.61} & \underline{65.63} & \textbf{68.51} & 59.67 & 70.66 & \underline{52.69} & \underline{62.48} \\
 & w/o Sampling Evaluation & 58.23 & 60.39 & 64.11 & 64.57 & 61.00 & \underline{71.04} & 50.11 & 61.35
 \\
 & w/o Selection & 57.47 & 60.93 & 64.66 & 60.28 & \underline{61.08} & 69.43 & 47.78 & 60.23 \\
\cmidrule(lr){2-10}
 & \textbf{SaM (Ours)} & \textbf{60.01} & \textbf{61.99} & \textbf{65.93} & \underline{67.05} & \textbf{62.41} & \textbf{71.65} & \textbf{52.90} & \textbf{63.13} \\
\bottomrule
\end{tabular}
}
\caption{Ablation studies. Removing certain components typically results in performance degradation, confirming their significance. The best results are in bold,
and the suboptimal ones are underlined.}
\label{tab:ablation_studies}
\end{table*}

\subsection{Ablation Studies}

We conduct ablation studies to validate the effectiveness of our design, as shown in Table~\ref{tab:ablation_studies}:  
(1) \textbf{w/o Merging}, which directly uses the best expert model.  
(2) \textbf{w/o Domain Similarity}, which selects experts solely based on Sampling Evaluation.  
(3) \textbf{w/o Sampling Evaluation}, which selects experts based on Domain Similarity.  
(4) \textbf{w/o Selection}, which merges all experts without selection.  
Results demonstrate that the merged model consistently outperforms the best individual expert, even when the selected models are not necessarily optimal. 
Overall, both expert selection strategies are effective and complementary, yielding the best results when combined. However, in some cases, using a single selection strategy or none at all yields better results, likely due to expert redundancy or insufficiency, as we fix the number of selected experts to $k=3$ in our experiments. 
Further analysis of $k$ are presented in Section~\ref{exp:expert_num}

\begin{table}[t]
\centering
\resizebox{\linewidth}{!}{
\begin{tabular}{lcccc}
\toprule
 & \textbf{Mode1} & \textbf{Mode2} & \textbf{Mode3} & \textbf{SaM (Ours)} \\
\midrule
\textbf{AI} & \underline{60.36} & 58.19 & \textbf{61.31} & 60.01 \\
\textbf{Literature} & 56.86 & \textbf{62.68} & 58.90 & \underline{61.99} \\
\textbf{Music} & 63.78 & \underline{66.50} & \textbf{66.85} & 65.93 \\
\textbf{Politics} & 66.05 & \textbf{68.27} & 61.30 & \underline{67.05} \\
\textbf{Science} & 58.37 & 61.73 & \textbf{63.12} & \underline{62.41} \\
\textbf{Movie} & 70.66 & \underline{70.66} & 70.50 & \textbf{71.65} \\
\textbf{Restaurant} & 52.82 & \underline{52.85} & 52.23 & \textbf{52.90} \\
\midrule
\textbf{Average} & 61.27 & \underline{62.98} & 62.03 & \textbf{63.13} \\
\bottomrule
\end{tabular}
}
\caption{Merging into a single task model (based on Qwen). ``Mode\textit{i}'' denotes the method used to further extract a final set from two expert sets for model merging.}
\label{tab:merge_single_model}
\vspace{-2mm}
\end{table}
\subsection{Merging into a Single Task Model}
\label{exp:merge_single_model}

Since our approach employs two task-specific models, the inference cost is doubled. To mitigate this, we derive a single set from the two selected expert sets, reducing the number of task models to one. We propose and evaluate three strategies, with results in Table~\ref{tab:merge_single_model}:  
(1) \textbf{Mode1} leverages the intersection of the two selected expert sets.  
(2) \textbf{Mode2} normalizes the evaluation metrics across both selection strategies (e.g., the domain similarity scores and F1-scores on sampled data points) to a common scale and selects the top three experts. 
(3) \textbf{Mode3} takes the union of the two expert sets while limiting the total number of selected experts to three.  
Experimental results show that Mode2 and Mode3 achieve comparable performance to us, making them effective alternatives without increasing inference costs. 
We refer to these as the economic versions of our framework (SaM$_{eco}$).

\begin{figure}[t]
  \centering
  \includegraphics[width=1.0\linewidth]{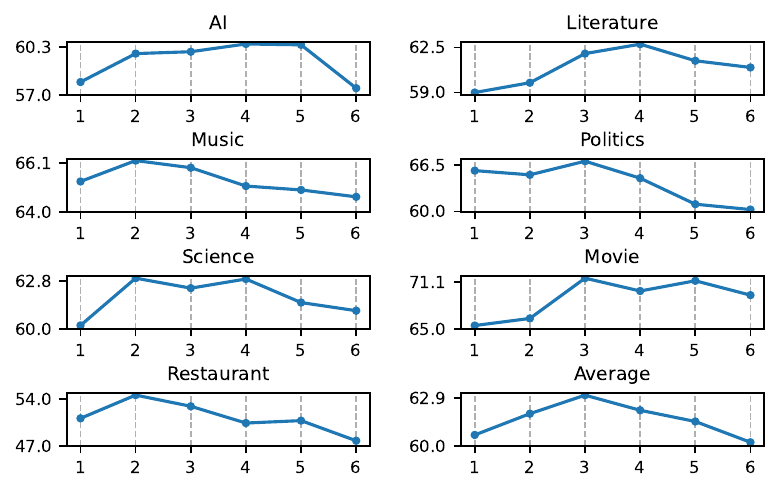}
  \caption{Performance changes with the number of experts. The horizontal axis is the number of experts for merging, and the vertical denotes the F1 scores.}
  \label{fig:expert_num_analysis}
  \vspace{-2mm}
\end{figure}
\subsection{Numbers of Experts for Merging}
\label{exp:expert_num}

This section analyzes the impact of the number of merged models, $k$, with results shown in Figure~\ref{fig:expert_num_analysis}. Here, $k=1$ denotes the performance of the best individual expert, while $k=6$ corresponds to merging all expert models. The optimal $k$ varies across target domains, typically ranging from 2 to 4. We set $k=3$ as it yields the best average performance. Notably, this corresponds to the merging algorithm as well. We adopt the Ties-Merging technique, where selecting $2\sim4$ models for merging is a commonly used configuration.

\begin{table*}[t]
\centering
\resizebox{0.95\textwidth}{!}{
\begin{tabular}{lcccccccc}
\toprule
Source Domains & \textbf{AI} & \textbf{Literature} & \textbf{Music} & \textbf{Politics} & \textbf{Science} & \textbf{Movie} & \textbf{Restaurant} & \textbf{Average} \\
\midrule
\textbf{Three} & 56.42 & 63.01 & 71.69 & 72.24 & 61.17 & 69.91 & 50.62 & 63.58 \\
\midrule
\textbf{Six} & 60.98 & 66.93 & 73.53 & 74.47 & 62.60 & 72.17 & 52.99 & 66.24 \\
\midrule
\textbf{Nine} & 59.90 & 66.93 & 73.53 & 73.61 & 62.78 & 72.17 & 53.84 & 66.11 \\
\bottomrule
\end{tabular}
}
\caption{Performance under different number of source domains. Using six source domains yields better results than three, but increasing to nine provides no further gains.}
\label{tab:num_of_source}
\vspace{-1mm}
\end{table*}

\begin{table*}[t]
\centering
\resizebox{0.95\textwidth}{!}{
\begin{tabular}{lcccccccc}
\toprule
 & \textbf{AI} & \textbf{Literature} & \textbf{Music} & \textbf{Politics} & \textbf{Science} & \textbf{Movie} & \textbf{Restaurant} & \textbf{Average} \\
\midrule
\textbf{Fully-trained} & 70.40 & 71.81 & 75.44 & 79.41 & 50.71 & 58.78 & 69.33 & 67.98 \\
\midrule
\textbf{Sam(Ours)} & 83.45 & 71.81 & 77.11 & 81.56 & 77.14 & 78.01 & 73.33 & 77.49 \\
\bottomrule
\end{tabular}
}
\caption{Performance under extreme scenarios with only one test sample (average of five trials). Our method still functions properly and adapts better than the fully trained single-model approach.}
\label{tab:one_test_sample}
\vspace{-2mm}
\end{table*}

\subsection{Number of Source Domains}

We set six source domains in our experiments. This section presents a preliminary analysis of how the number of source domains affects performance, as shown in Table~\ref{tab:num_of_source}. While setting more source domains increases the diversity of candidate models during selection, it does not always lead to better results. The performance relies on the similarity between source and target domains, including conceptual relevance, data distribution, and overlap in entity types. Therefore, it's important to balance the number of source domains, their similarity to target domains, and the overall complexity.

\subsection{Limited Target Resources Scenarios}

The model selection process leverages target-domain raw texts for domain similarity calculation and sampling evaluation, typically a few hundred samples for the former and 10 for the latter. To test our method under more constrained settings, we simulate an extreme case with only one target-domain instance. As shown in Table~\ref{tab:one_test_sample}, our method remains effective. However, very small sample pools can undermine the stability of expert selection, especially for the strategy using domain similarity. In such cases, we recommend using data augmentation to expand the sample pool.

\begin{table}[t]
\footnotesize
\centering
\resizebox{0.97\linewidth}{!}{
\begin{tabular}{lcccc}
\toprule
 & \textbf{Mode1} & \textbf{Mode2} & \textbf{SaM (Ours)} \\
\midrule
\textbf{AI} & \textbf{60.07} & 59.33 & 60.01 \\
\textbf{Literature} & \textbf{62.00} & 62.03 & 61.99 \\
\textbf{Music} & \textbf{66.35} & 65.17 & 65.93 \\
\textbf{Politics} & 65.54 & 66.45 & \textbf{67.05} \\
\textbf{Science} & 62.27 & 62.15 & \textbf{62.41} \\
\textbf{Movie} & 71.20 & 71.48 & \textbf{71.65} \\
\textbf{Restaurant} & \textbf{53.84} & 52.84 & 52.90 \\
\midrule
\textbf{Average} & 63.04 & 62.78 & \textbf{63.13} \\
\bottomrule
\end{tabular}
}
\caption{Weighting experts for merging. ``Mode\textit{i}'' denotes the method used to weight model parameters.}
\label{tab:weghted_merge}
\vspace{-2mm}
\end{table}

\subsection{Weighting Experts for Merging}
\label{weighting_experts}

We employ two metrics, \textit{domain similarity} and \textit{sampling evaluation}, to select expert models. These metrics reflect the importance of experts. Our framework does not consider the importance of experts but instead assigns equal weight to all expert models.
To investigate the impact, we propose two simple weighting strategies, as shown in Table~\ref{tab:weghted_merge}:
(1) \textbf{Mode1} empirically assigns weights $(1.5, 1.0, 0.5)$ to the top three selected experts.
(2) \textbf{Mode2} uses the middle-ranked metric value as a normalization factor to scale the three experts’ metrics for weighting. 
Experimental results suggest that weighting has the potential to improve performance (Mode1), aligning with intuitive expectations. 
However, excessive reliance on heuristics may not always be justified. For example, while Mode2 theoretically provides a more precise weighting based on expert importance, it underperforms compared to Mode1 and Ours.

\subsection{Finer Adaptation for Target Domains}
\label{exp:finer_adaptation}

\begin{figure}[t]
  \centering
  \includegraphics[width=1.0\linewidth]{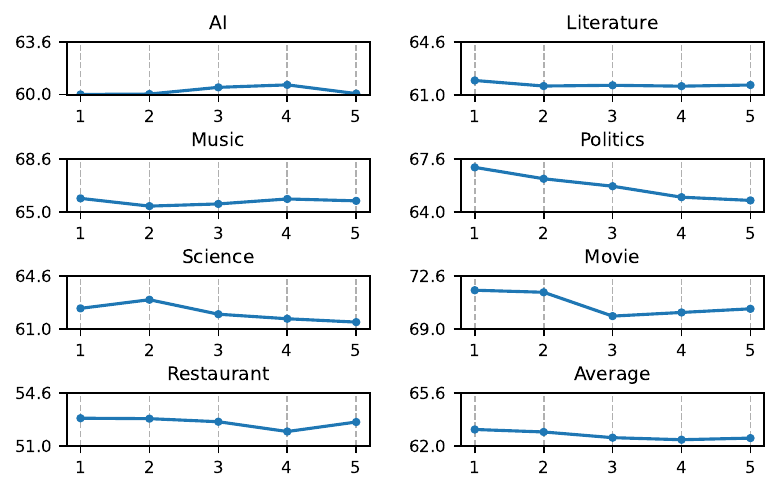}
  \caption{Performance changes with the number of data splits. The horizontal axis is the number of splits, and the vertical denotes the entity-level F1 scores.}
  \label{fig:cluster_analysis_1}
\end{figure}

Beyond the weighted merging strategy in Section~\ref{weighting_experts}, another potential approach for improvements is adopting finer adaptation. Specifically, we apply clustering to divide the target domain into multiple splits, selecting and merging expert models separately for each. 
However, as shown in Figure~\ref{fig:cluster_analysis_1}, this finer adaptation does not enhance performance. Instead, it results in an overall decline.

\begin{figure}[t]
  \centering
  \includegraphics[width=1.0\linewidth]{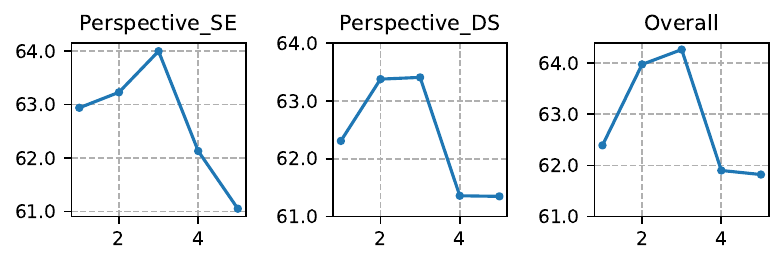}
  \caption{Performance changes with the number of data splits. We aggregate data from all domains into a unified dataset and subsequently conduct the split analysis.}
  \label{fig:cluster_analysis_2}
  \vspace{-2mm}
\end{figure}

To investigate this further, we aggregate data from all seven domains into a single dataset and conduct experiments. As shown in Figure~\ref{fig:cluster_analysis_2}, clustering improves performance across various strategies, including Sampling Evaluation (SE), Domain Similarity (DS), and their combination.  
Notably, despite the dataset spanning seven domains, the best performance is achieved when clustering divides it into only two or three groups. This is intuitively reasonable, as these seven domains originate from two broader datasets (CrossNER and MIT).
The above analysis indicates that while finer adaptation to the target domain may bring improvements, excessive refinement without constraints may be counterproductive. For example, treating each data point as a distinct domain might seem optimal in theory but leads to poor performance in practice. 

\begin{table}[t]
\centering
\resizebox{0.96\linewidth}{!}{
\begin{tabular}{lcccc}
\toprule
 & \textbf{dare-linear} & \textbf{ties} & \textbf{dare-ties} \\
\midrule
\textbf{AI} & 54.67 & \textbf{60.01} & 52.57 \\
\textbf{Literature} & 56.22 & \textbf{61.99} & 55.81 \\
\textbf{Music} & \textbf{66.31} & 65.93 & 66.22 \\
\textbf{Politics} & \textbf{69.11} & 67.05 & 66.42 \\
\textbf{Science} & 62.01 & \textbf{62.41} & 61.07 \\
\textbf{Movie} & 68.31 & \textbf{71.65} & 65.12 \\
\textbf{Restaurant} & 52.12 & \textbf{52.90} & 51.40 \\
\midrule
\textbf{Average} & 61.25 & \textbf{63.13} & 59.80 \\
\bottomrule
\end{tabular}
}
\caption{Comparison of different merging techniques. We compare linear, dare, ties, and some combinations.}
\label{tab:merge_technique}
\end{table}
\subsection{Analysis of Merging Technique}

In addition to the Ties-Merging algorithm (\textbf{ties}) we employed, this section analyzes alternative merging strategies, including \textbf{linear}, \textbf{dare}~\cite{yu2024dare}, and their combinations. 
As shown in Table~\ref{tab:merge_technique}, dare and ties are effective.
We do not present the performance of linear strategy, as this strategy leads to significant degradation. Specifically, models merged via the linear approach still produce some meaningful content but almost lose the ability to generate structured outputs, which is crucial for NER and other IE tasks. Consequently, their performance is bad, though minor improvements can be achieved through extensive post-processing on model outputs. However, combining dare with linear yields improved results.
Both dare and ties address the issue of parameter redundancy for merging. 
These findings suggest that handling parameter redundancy is crucial for NER and similar structured output tasks. Additionally, the relatively weaker performance of the dare-ties combination may stem from excessive redundancy reduction, which could compromise useful capabilities of models.

\begin{table}[t]
\centering
\resizebox{\linewidth}{!}{
\begin{tabular}{lcccccc}
\toprule
 & \multicolumn{3}{c}{\textbf{Llama}} & \multicolumn{3}{c}{\textbf{Qwen}} \\
\cmidrule(lr){2-4} \cmidrule(lr){5-7}
 & T-SC & E-SC & Ours & T-SC & E-SC & Ours \\
\cmidrule(lr){1-1} \cmidrule(lr){2-4} \cmidrule(lr){5-7}
\textbf{AI} & 54.20 & 60.31 & 60.98 & 51.47 & 57.36 & 60.01 \\
\textbf{Literature} & 64.13 & 58.98 & 66.93 & 54.61 & 56.08 & 61.99 \\
\textbf{Music} & 70.01 & 71.06 & 73.53 & 61.58 & 68.01 & 65.93 \\
\textbf{Politics} & 63.58 & 66.27 & 74.47 & 55.76 & 65.99 & 67.05 \\
\textbf{Science} & 59.01 & 60.03 & 62.60 & 58.73 & 60.99 & 62.41 \\
\textbf{Movie} & 64.72 & 67.78 & 72.17 & 66.58 & 67.86 & 71.65 \\
\textbf{Restaurant} & 50.21 & 50.31 & 52.99 & 52.50 & 50.43 & 52.90 \\
\cmidrule(lr){1-1} \cmidrule(lr){2-4} \cmidrule(lr){5-7}
\textbf{Average} & 60.84 & 62.11 & \textbf{66.24} & 57.32 & 60.96 & \textbf{63.13} \\
\bottomrule
\end{tabular}
}
\caption{Comparison with self-consistency (SC) methods. T-SC employs a fully trained model to generate multiple outputs by adjusting the \underline{T}emperature hyperparameter to the ensemble, while E-SC ensemble outputs from different \underline{E}xpert models.}
\label{tab:compare_sc}
\end{table}
\subsection{Comparing with Self-Consistency}

Considering that we trained multiple expert models, an intuitive approach is self-consistency (SC)~\citep{wang2023sc}, which ensembles multiple outputs through voting. Results are shown in Table~\ref{tab:compare_sc}, where \textbf{T-SC} employs a full-trained model to generate multiple outputs by adjusting the \textbf{T}emperature hyperparameter to ensemble, while \textbf{E-SC} ensemble outputs from different \textbf{E}xperts (though this slightly deviates from the strict definition of ``self''-consistency, we refer to "SC" for simplicity). For our NER task, traditional SC (T-SC) shows limited improvement, while E-SC offers more significant gains due to greater output diversity. However, both SC strategies perform worse than our method and require higher inference costs.

\subsection{Framework Generalization}
\paragraph{Non-strict Domain}
In our experiments, each expert corresponds to a domain with real-world significance, such as news or law. To explore a more flexible scenario, we further test on non-strict domains. Specifically, we cluster each domain dataset into five subsets, and then sample one subset from each domain to create five new datasets, which are used to train five new expert models.
The results in Table~\ref{tab:new_domain_experts} indicate that our framework remains effective.
Notably, the experts in this setting are no longer tied to specific domains but function as general-domain models while retaining diverse capabilities. This suggests that the key requirement is a set of experts with complementary strengths, which can enhance overall performance through mutual reinforcement.

\paragraph{Multilingual Scenarios}
We extend our framework to multilingual scenarios and conduct preliminary experiments on six languages from the WikiANN dataset~\cite{pan2017wikiann}, including German (de), English (en), Spanish (es), Dutch (nl), Russian (ru), and Chinese (zh). We train a model for each language. When evaluating a target language, the model trained on that language is not used. The results in Table~\ref{tab:linguistic_experts} provide several initial evidence that our framework has the potential to extend to multilingual scenarios.

\section{Future Work}
\paragraph{Unified IE and Other Tasks}
We conduct experimental analyses with NER as a case study. Some prior studies train a unified model for multiple IE tasks, including NER, relation extraction (RE), event extraction (EE), etc. Our framework can be naturally extended to a broader IE setting by incorporating additional IE data to train IE experts.
Additionally, our method extends beyond these tasks to a wide range of applications.

\paragraph{Detailed and Complete Design}
Our extended experiments investigated several optimization strategies. Further improvements could be realized by:
(1) Incorporating more task-specific designs. For instance, alongside domain‐level similarity, we could also leverage entity‐type similarity when selecting source models.
(2) Dynamically determining the number of merged models, $k$. Model merging may be unnecessary for certain target domains. As shown in Appendix~\ref{appendix_B}, when the target and source domains coincide, the single corresponding model already delivers optimal results. Section~\ref{exp:expert_num} demonstrates that the best choice of $k$ varies across different target domains. Additionally, our framework is agnostic to the model architecture and readily extends to other model types other than the Llama and Qwen LLM families.

\begin{table}[t]
\centering
\resizebox{\linewidth}{!}{
\begin{tabular}{lcccccc}
\toprule
 & \multicolumn{5}{c}{\textbf{Expert Model}} & \multirow{2}{*}{\textbf{Ours}} \\
\cmidrule(lr){2-6}
 & E1 & E2 & E3 & E4 & E5 &  \\
\midrule
\textbf{AI} & \underline{56.98} & 57.98 & 56.26 & 48.51 & 55.46 & \textbf{60.88} \\
\textbf{Literature} & 65.38 & \underline{65.45} & 55.30 & 56.27 & 62.20 & \textbf{67.03} \\
\textbf{Music} & 62.38 & \underline{66.00} & 63.63 & 63.88 & 65.17 & \textbf{67.42} \\
\textbf{Politics} & 60.94 & \textbf{66.49} & 63.52 & 57.54 & 58.50 & \underline{66.09} \\
\textbf{Science} & 62.56 & 63.31 & 61.38 & 58.23 & \underline{65.72} & \textbf{66.75} \\
\textbf{Movie} & \textbf{68.75} & 61.97 & 66.58 & 64.68 & 65.02 & \underline{67.98} \\
\textbf{Restaurant} & 44.06 & 34.94 & 49.57 & \underline{50.05} & 43.35 & \textbf{52.66} \\
\midrule
\textbf{Average} & \underline{60.15} & 59.45 & 59.46 & 57.02 & 59.35 & \textbf{64.12} \\
\bottomrule
\end{tabular}
}
\caption{Analysis of experts for non-strict domains. We employ clustering to build five source domains and train expert models.}
\label{tab:new_domain_experts}
\end{table}

\begin{table}[t]
\centering
\resizebox{\linewidth}{!}{
\begin{tabular}{lccccccc}
\toprule
 & \multicolumn{6}{c}{\textbf{Expert Model}} & \multirow{2}{*}{\textbf{Ours}} \\
\cmidrule(lr){2-7}
 & de & en & es & nl & ru & zh \\
\midrule
\textbf{de} & -- & 80.69 & 80.75 & 81.00 & \underline{81.74} & 74.99 & \textbf{82.42} \\
\textbf{en} & 72.29 & -- & \underline{77.10} & \textbf{77.51} & 73.08 & 69.18 & 76.84 \\
\textbf{es} & 86.09 & 87.28 & -- & \textbf{91.30} & 88.46 & 78.85 & \underline{89.67} \\
\textbf{nl} & 84.31 & \underline{86.75} & 85.02 & -- & 84.79 & 81.41 & \textbf{87.59} \\
\textbf{ru} & 75.91 & 75.68 & 74.44 & \underline{76.93} & -- & 61.29 & \textbf{78.16} \\
\textbf{zh} & 49.07 & 49.18 & 46.56 & 48.21 & \underline{49.93} & -- & \textbf{51.37} \\
\midrule
\textbf{Avg} & 73.53 & \underline{75.92} & 72.77 & 74.99 & 75.60 & 73.14 & \textbf{77.68} \\
\bottomrule
\end{tabular}
}
\caption{Analysis of experts for different languages. When evaluating a target language, the model trained on that language is not used.}
\label{tab:linguistic_experts}
\vspace{-2mm}
\end{table}

\section{Conclusion}
We propose the Select and Merging (SaM) framework for NER, which dynamically selects valuable domain expert models for the target domain and employs model merging to obtain the task-specific model. 
Compared to prior studies, we possess superior adaptability and scalability.  
Experimental results demonstrate the effectiveness of our framework. 
Extensive analysis further provides insights into potential improvements, practical experience, and broader extensions of our approach.

\section*{Limitations}
\label{limitations}
We acknowledge the following limitations of our work:
(1) Maintaining multiple expert models introduces some additional storage overhead, despite the use of LoRA.  
(2) For domain similarity calculation and clustering analysis, we simply employed a widely used encoder model from the HuggingFace repository to obtain text embeddings. Further optimization is possible.
(3) Our analysis is limited to Named Entity Recognition (NER). Further experiments are needed for other IE tasks.

\section*{Acknowledgements}
This work was supported in part by the National Natural Science Foundation of China under Grant No.62276110,  No.62172039. The authors would also like to thank anonymous reviewers for their comments on improving the quality of this paper. 

\bibliography{main}

\clearpage
\appendix

\section{Experimental Details}
\label{apd:training_details}
\paragraph{Training Data}

\begin{figure}[t]
  \centering
  \includegraphics[width=1.0\linewidth]{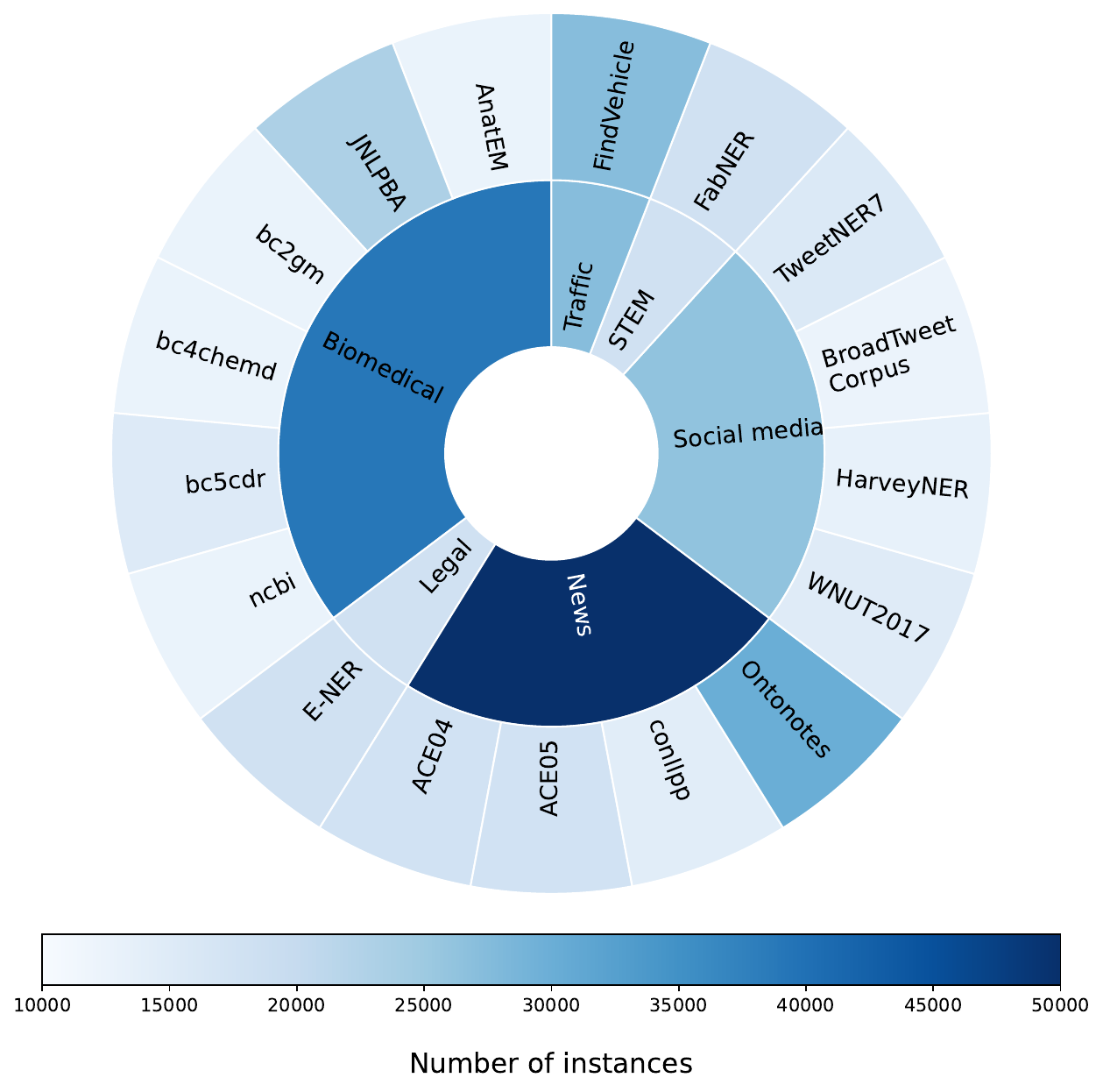}
  \caption{Distribution of the training data.}
  \label{fig:traing_data}
\end{figure}

\begin{figure*}[ht]
  \centering
  \includegraphics[width=1.0\linewidth]{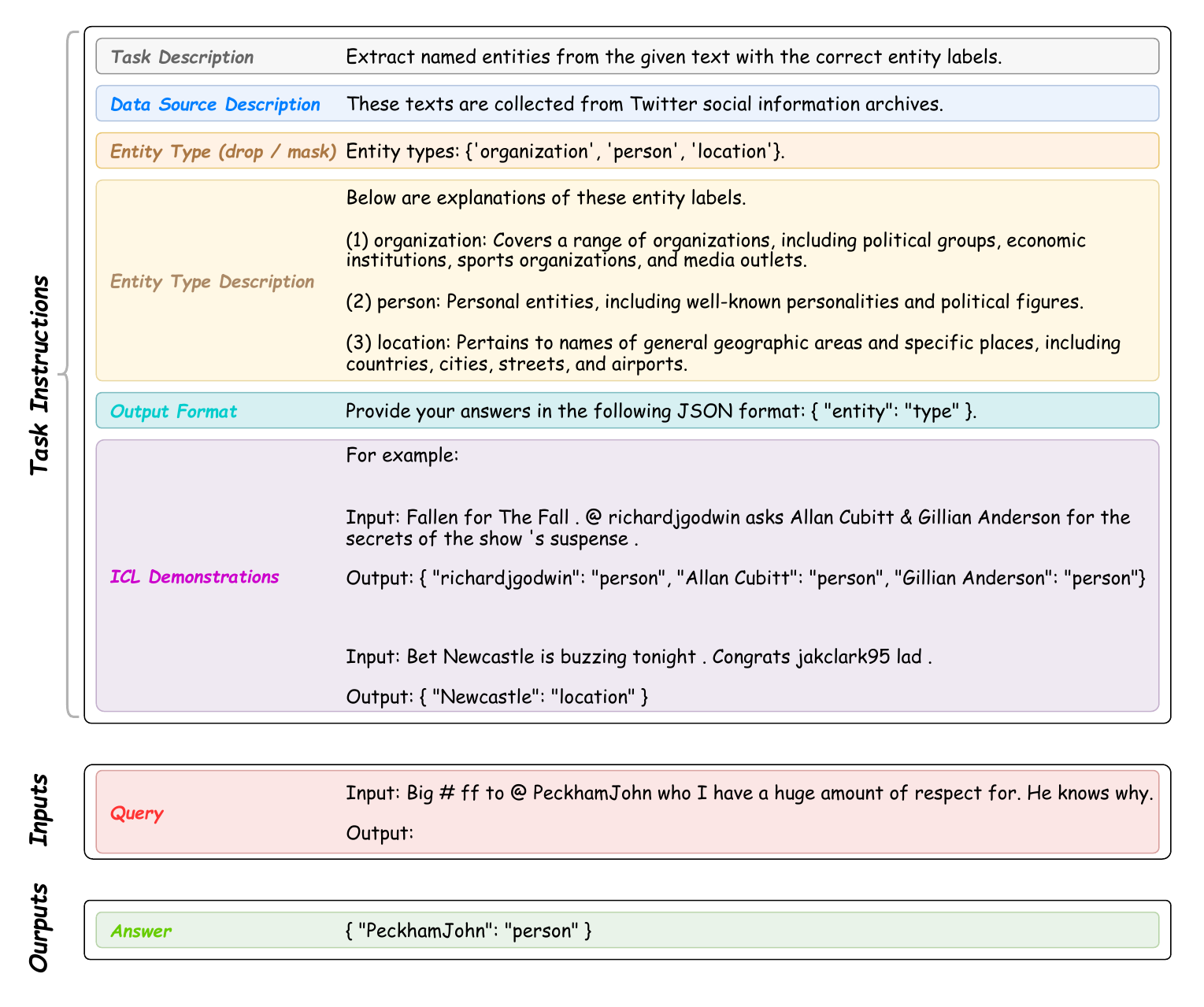}
  \caption{Formatted training data example. The example consists of task instructions, inputs, and outputs for training. 
  For the task instructions, the Data Source Description, Entity Type drop/mask, and ICL Demonstrations are optional, with details in section~\ref{method_data_collection}. We adopt three output formats: JSON, enumeration, and natural language descriptions. The output format of ICL Demonstrations and Answers should be consistent with the specified.}
  \label{fig:instruction_case}
\end{figure*}

\begin{figure*}[ht]
  \centering
  \includegraphics[width=1.0\linewidth]{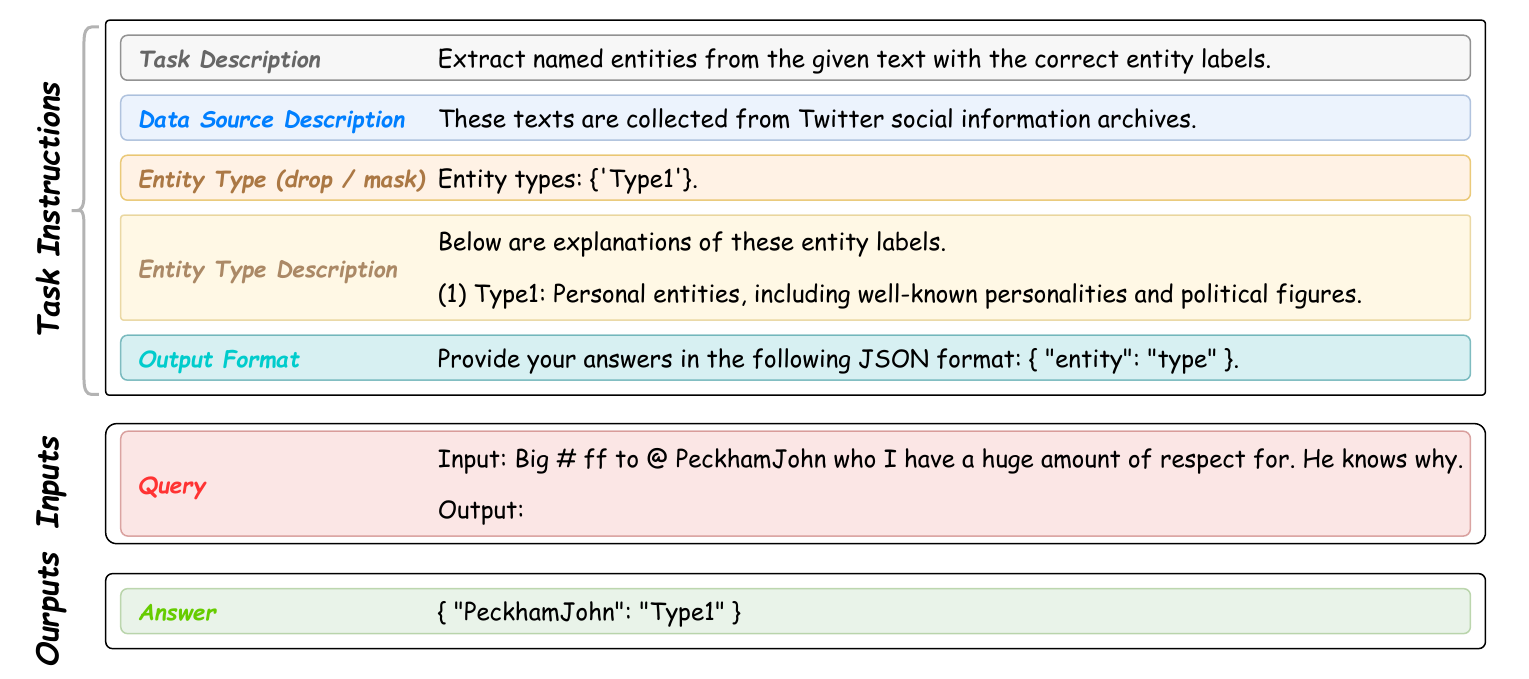}
  \caption{Another example of our formatted training data. The instance here is the same as that of Figure~\ref{fig:instruction_case}, adopting the entity type drop and mask processing methods.}
  \label{fig:instruction_case_2}
\end{figure*}

We collected 17 widely used NER datasets and categorized them into six domains based on their data sources. Table~\ref{tab:training_data} presents detailed statistics for each dataset. As described in Section~\ref{method_data_collection}, we sample these datasets based on the proportion of their entity types while ensuring that the total data volume per domain remains between 10,000 and 50,000. The distribution of the training data across domains is illustrated in Figure~\ref{fig:traing_data}, and the exact number of sampled instances per dataset is listed in the ``\#Sampled'' column of Table~\ref{tab:training_data}.
Figure~\ref{fig:instruction_case} and~\ref{fig:instruction_case_2} show examples of the formatted training data we constructed.

\paragraph{Implementation Details}
We adopt the Qwen2.5-7B ~\footnote{\url{https://huggingface.co/Qwen/Qwen2.5-7B}} 
and 
Llama3.1-8B ~\footnote{\url{https://huggingface.co/meta-llama/Llama-3.1-8B}}
as foundations and train expert models using LoRA~\citep{hu2022lora}. 
The LoRA rank is set to 32, with three training epochs, a batch size of 16, a learning rate of $2e-5$, and a warmup ratio of 0.05. During LoRA training, all linear layers are activated.
We set the temperature to 0 for LLMs when inference.
All experiments are conducted on one NVIDIA 4090 GPU.

\section{Data Merging or Model Merging}
\label{appendix_B}

We use \textbf{``Data Merging''} to denote the prior approaches of training a unified model by integrating data from multiple domains and \textbf{``Model Merging''} to denote synthesizing a new model by merging the parameters of expert models. This section presents a preliminary analysis that motivates the adoption of the model merging strategy. 
Specifically, we compare three types of models: (1) Experts, which are trained on single-domain data. (2) Data Merging, which is trained on a mixture of \textbf{all} domain data. (3) Model Merging, which is obtained by merging the parameters of \textbf{all} expert models.

We conduct evaluations on both in-domain and out-of-domain settings, with the results presented in Tables~\ref{tab:in_domain} and~\ref{tab:out_of_domain}. Our key observations are as follows:  
\textbf{(1)} Data Merging consistently yields suboptimal performance in both settings. \textbf{(2)} In in-domain tasks, Model Merging also performs suboptimally and is inferior to Data Merging. \textbf{(3)} In out-of-domain tasks, Model Merging generally achieves the best performance.

The aforementioned observation is our primary motivation for introducing model merging. For in-domain tasks, directly utilizing the corresponding expert model is sufficient. For out-of-domain tasks, merging expert models improves generalization. However, experimental results indicate that in some cases, a single expert model performs better, especially for in-domain tasks where merging may be unnecessary. Combined with Section~\ref{exp:expert_num}, the number of selected experts, $k$, could be dynamically determined, with $k=1$ being a valid consideration. This is a promising direction for further enhancing the adaptability of our framework.

\begin{table*}[ht]
\centering
\resizebox{\textwidth}{!}{
\begin{tabular}{cccrrrr}
\toprule
\textbf{Domain} & \textbf{Dataset} & \textbf{\#Type} & \textbf{\#Train} & \textbf{\#Dev} & \textbf{\#Test} & \textbf{\#Sampled} \\
\midrule
\multirow{6}{*}{\textbf{Biomedical}} & AnatEM~\citep{pyysalo2014AnatEM} & 1 & 5,861 & 2,118 & 3,830 & 3,297  \\
 & JNLPBA~\citep{collier2004JNLPBA} & 5 & 16,691 & 1,855 & 3,856 & 16,487 \\
 & bc2gm~\citep{smith2008bc2gm} & 1 & 12,500 & 2,500 & 5,000 & 3,297 \\
 & bc4chemd~\citep{krallinger2015bc4chemd} & 1 & 30,682 & 30,639 & 26,364 & 3,297 \\
 & bc5cdr~\citep{li2016bc5cdr} & 2 & 4,560 & 4,581 & 4,797 & 6,594 \\
 & ncbi~\citep{dougan2014ncbi} & 1 & 5,432 & 923 & 940 & 3,297 \\
\midrule
\textbf{Law} & E-NER~\citep{au2022E-NER} & 7 & 9,313 & 1,164 & 1,165 & 10,000 \\
\midrule
\multirow{4}{*}{\textbf{News}} & ACE04~\citep{walker2006ACE04} & 7 & 6,202 & 745 & 812 & 9,722 \\
 & ACE05~\citep{walker2006ACE05} & 7 & 7,299 & 971 & 1,060 & 9,722 \\
 & conllpp~\citep{wang2019conllpp} & 4 & 14,041 & 3,250 & 3,452 & 5,555 \\
 & OntoNotes~\citep{pradhan2013OntoNotes} & 18 & 59,924 & 8,528 & 8,262 & 25,000 \\
\midrule
\multirow{4}{*}{\hspace{1em}\textbf{Social media}\hspace{1em}} & WNUT2017~\citep{derczynski2017WNUT2017} & 6 & 3,394 & 1,009 & 1,287 & 6,094 \\
 & HarveyNER~\citep{chen2022HarveyNER} & 4 & 3,967 & 1,301 & 1,303 & 4,063 \\
 & BroadTweetCorpus~\citep{derczynski2016BroadTweetCorpus} & 3 & 5,334 & 2,001 & 2,000 & 3,047 \\
 & TweetNER7~\citep{ushio2022TweetNER7} & 7 & 7,111 & 886 & 576 & 7,110 \\
\midrule
\textbf{STEM} & FabNER~\citep{kumar2022FabNER} & 11 & 9,435 & 2,182 & 2,064 & 10,000 \\
\midrule
\textbf{Traffic} & FindVehicle~\citep{guan2024FindVehicle} & 8 & 21,565 & 20,777 & 20,777 & 21,565 \\
\bottomrule
\end{tabular}
}
\caption{Statistics of raw training data and the number of sampled instances for training.}
\label{tab:training_data}
\end{table*}

\begin{table*}[t]
\centering
\resizebox{0.97\textwidth}{!}{
\begin{tabular}{llcccccc}
\toprule
 & & \textbf{Biomedical} & \textbf{Legal} & \textbf{\hspace{1em}News\hspace{0.5em}} & \textbf{Socia media} & \textbf{STEM} & \textbf{Traffic} \\
\midrule
\multirow{6}{*}{\textbf{\hspace{1em}Experts\hspace{1em}}} & Biomedical & \textbf{82.57} & 40.02 & 36.61 & 48.09 & 27.27 & 21.05 \\
 & Legal & 40.40 & \textbf{84.79} & 42.36 & 46.90 & 17.92 & 32.48 \\
 & News & 53.83 & 48.40 & \textbf{85.94} & 48.28 & 23.53 & 41.84 \\
 & Social media & 53.11 & 41.31 & 42.06 & \textbf{66.57} & 24.17 & 22.96 \\
 & STEM & 28.22 & 16.26 & 22.81 & 23.53 & \textbf{76.99} & 24.96 \\
 & Traffic & 42.14 & 28.23 & 32.67 & 42.01 & 20.23 & \textbf{99.96} \\
\midrule
\multicolumn{2}{c}{\textbf{Data Merging}} & \underline{80.53} & \underline{81.87} & \underline{84.97} & \underline{60.27} & \underline{77.40} & \underline{98.91} \\
\midrule
\multicolumn{2}{c}{\textbf{Model Merging}} & 67.57 & 64.13 & 55.47 & 56.25 & 31.80 & 45.29\\
\bottomrule
\end{tabular}
}
\caption{In-domain performance of expert models, full data trained model (Data Merging), and model obtained from merging all expert models (Model Merging).}
\label{tab:in_domain}
\end{table*}

\begin{table*}[ht!]
\centering
\resizebox{0.97\textwidth}{!}{
\begin{tabular}{llccccccc}
\toprule
 & & \textbf{AI} & \textbf{Literature} & \textbf{Music} & \textbf{Politics} & \textbf{Science} & \textbf{Movie} & \textbf{Restaurant} \\
\midrule
\multirow{6}{*}{\textbf{\hspace{1em}Experts\hspace{1em}}} & Biomedical & \underline{56.90} & \underline{59.01} & 63.98 & \textbf{65.73} & \underline{60.21} & 57.96 & 39.41 \\
 & Legal & 39.85 & 56.88 & \underline{64.33} & 62.72 & 55.92 & 63.26 & 47.31 \\
 & News & 41.77 & 41.03 & 54.74 & 40.36 & 49.66 & 62.13 & 39.28 \\
 & Social media & 53.84 & 55.83 & 61.14 & 60.50 & 57.24 & 59.29 & 42.63 \\
 & STEM & 41.01 & 41.35 & 43.35 & 45.41 & 37.38 & 46.61 & 26.78 \\
 & Traffic & 49.97 & 50.82 & 62.20 & \underline{64.09} & 53.25 & 65.45 & \underline{52.11} \\
\midrule
\multicolumn{2}{c}{\textbf{Data Merging}} & 51.38 & 53.46 & 61.12 & 54.99 & 59.39 & \underline{66.66} & \textbf{52.69} \\
\midrule
\multicolumn{2}{c}{\textbf{Model Merging}} & \textbf{57.47} & \textbf{61.93} & \textbf{64.66} & 60.28 & \textbf{61.08} & \textbf{69.43} & 47.78 \\
\bottomrule
\end{tabular}
}
\caption{Our-of-domain performance of expert models, full data trained model (Data Merging), and model obtained from merging all expert models (Model Merging).}
\label{tab:out_of_domain}
\end{table*}

\begin{table*}[t]
\centering
\resizebox{0.97\linewidth}{!}{
\begin{tabular}{lccccccc}
\toprule
 & \multicolumn{5}{c}{\textbf{Unified Models}} & \multirow{2}{*}{\textbf{SaM}} & \multirow{2}{*}{\textbf{SaM$_{eco}$}} \\
\cmidrule(lr){2-6}
 & InstructUIE & UniNER & GoLLIE & B2NER & Ours & & \\
\midrule
\textbf{Training Instances} & 215.9K & 45.9K & 165.2K & 51.9K & 148.1K & 148.1K & 148.1K \\
\textbf{Inference Times} & 1$\times$ & 1$\times$ & 1$\times$ & 1$\times$ & 1$\times$ & 2$\times$ & 1$\times$ \\
\textbf{Storage (Normalized)} & 1 & 1 & 1 & 1 & 1 & 1+0.02n & 1+0.02n \\
\midrule
\textbf{Performance (Average)} & 47.84 & 61.79 & 58.39 & 64.57 & 60.94 & 66.24 & 65.49  \\
\bottomrule
\end{tabular}
}
\caption{Comparison of Resource Requirements. We compare unified models trained across multiple domains with our merging-based approach. SaM$_{eco}$ (economic) refers to integrating two task-specific models into a single one (details in Section~\ref{exp:merge_single_model}). For storage, we normalize the value by setting the model size to 1. Here, $n$ represents the number of experts. We achieve superior results with minimal additional overhead, particularly with our SaM$_{eco}$.}
\label{tab:cost}
\vspace{1.2mm}
\end{table*}

\begin{table*}[t]
\centering
\resizebox{0.97\linewidth}{!}{
\begin{tabular}{lcccccccc}
\toprule
 & \multicolumn{4}{c}{Llama} & \multicolumn{4}{c}{Qwen} \\
\cmidrule(lr){2-5} \cmidrule(lr){6-9}
 & \textbf{Mode1} & \textbf{Mode2} & \textbf{Mode3} & \textbf{SaM(Ours)} & \textbf{Mode1} & \textbf{Mode2} & \textbf{Mode3} & \textbf{SaM(Ours)} \\
\midrule
\textbf{AI} & \underline{60.36} & 58.19 & \textbf{61.31} & 60.01 & 54.73 & 55.62 & 57.90& \textbf{60.98}\\
\textbf{Literature} & 56.86 & \textbf{62.68} & 58.90 & \underline{61.99} & 60.16 & 59.87 & \underline{63.42} & \textbf{66.93} \\
\textbf{Music} & 63.78 & \underline{66.50} & \textbf{66.85} & 65.93 & 68.98 & \underline{72.89} & 72.42 & \textbf{73.53} \\
\textbf{Politics} & 66.05 & \textbf{68.27} & 61.30 & \underline{67.05} & 71.16 & \textbf{75.42} & 73.61 & \underline{74.47} \\
\textbf{Science} & 58.37 & 61.73 & \textbf{63.12} & \underline{62.41} & 62.19 & 59.58 & \textbf{62.78} & \underline{62.60} \\
\textbf{Movie} & 70.66 & \underline{70.66} & 70.50 & \textbf{71.65} & 67.07 & 68.91 & \textbf{73.44} & \underline{72.17} \\
\textbf{Restaurant} & 52.82 & \underline{52.85} & 52.23 & \textbf{52.90} & 51.69 & 51.69 & \textbf{54.84} & \underline{52.99} \\
\midrule
\textbf{Average} & 61.27 & \underline{62.98} & 62.03 & \textbf{63.13} & 62.28 & 63.43 & \underline{65.49} & \textbf{66.24} \\
\bottomrule
\end{tabular}
}
\caption{Merging into a single task model. Complete experimental results of Section~\ref{exp:merge_single_model}.
}
\label{tab:merge_single_model_apendix}
\end{table*}

\section{Analysis of Cost}
\label{appendix:cost}
\paragraph{Parameter Count and Storage Cost}
We assume \( H \) denotes the model dimension, \( r \) the rank of LoRA adapters, \( L \) the number of layers, \( n \) the number of domain-specific experts, and \( V \) the vocabulary size. For computational simplicity, we adopt a simplified Transformer architecture (e.g., omitting grouped-query attention mechanisms) in our base models (Llama3.1 and Qwen2.5). Since our LoRA implementation applies to all linear layers, each domain-specific expert requires \( 18HrL \) additional storage parameters. Consequently, storing \( n \) experts incurs a total overhead of \( n \times 18HrL \). The base model itself requires approximately \( (12H^2 + 13H)L + VH \) parameters. Given that \( nr \ll H \), the additional storage overhead remains negligible. During inference, we merge LoRA adapters into one or two task-specific models, achieving state-of-the-art performance with only equivalent or doubled storage costs compared to the base model.
\paragraph{Computation FLOPs Analysis}
Compared to multi-task full fine-tuning (FFT), our approach utilizes the same amount of training data to produce multiple LoRA models, resulting in identical computational costs during training. During inference, however, our method employs a single task-specific model—either derived from a single strategy or by integrating models from both strategies—which requires only one LoRA adapter. This achieves comparable inference costs to multi-task FFT while delivering superior performance. When leveraging models from both strategies simultaneously, our approach incurs twice the inference cost but further enhances performance, offering a flexible trade-off between efficiency and effectiveness.
Table~\ref{tab:cost} compares our approach with several recent works that train a unified model across multiple domains. We compare (1) the amount of training data (with instance-level data size provided for reference), (2) the number of inference rounds required for model prediction, and (3) storage space requirements (with model size as the reference unit). We also report the average performance.

\section{Experimental Supplements}

Section~\ref{exp:merge_single_model} extracts a single expert set for merging and presents the results based on Qwen. Here, we supplement the results of Llama in Table~\ref{tab:merge_single_model_apendix}.

Our framework is based on two perspectives: Domain Similarity (DS) and Sampling Evaluation (SE). The experimental section reports the overall performance of the framework. Here, we provide several additional experimental results regarding these two strategies, respectively.

\begin{figure*}[t]
  \centering
  \includegraphics[width=1.0\linewidth]{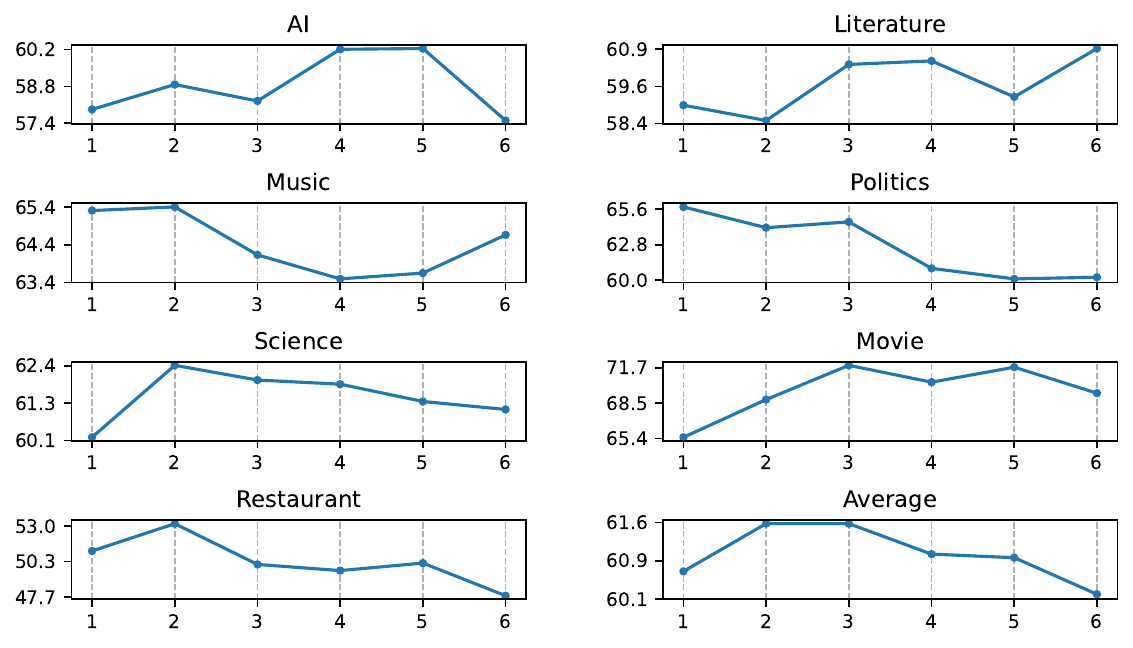}
  \caption{Performance changes with the number of expert models (denotes as $k$). The horizontal axis is the number of experts for merging, and the vertical denotes the entity-level F1 scores. \textbf{Only using Domain Similarity for expert selection}. It can be seen that the optimal $k$ varies across target domains, typically ranging from 2 to 4.}
  \label{fig:expert_num_ds}
\end{figure*}

\begin{figure*}[t]
  \centering
  \includegraphics[width=1.0\linewidth]{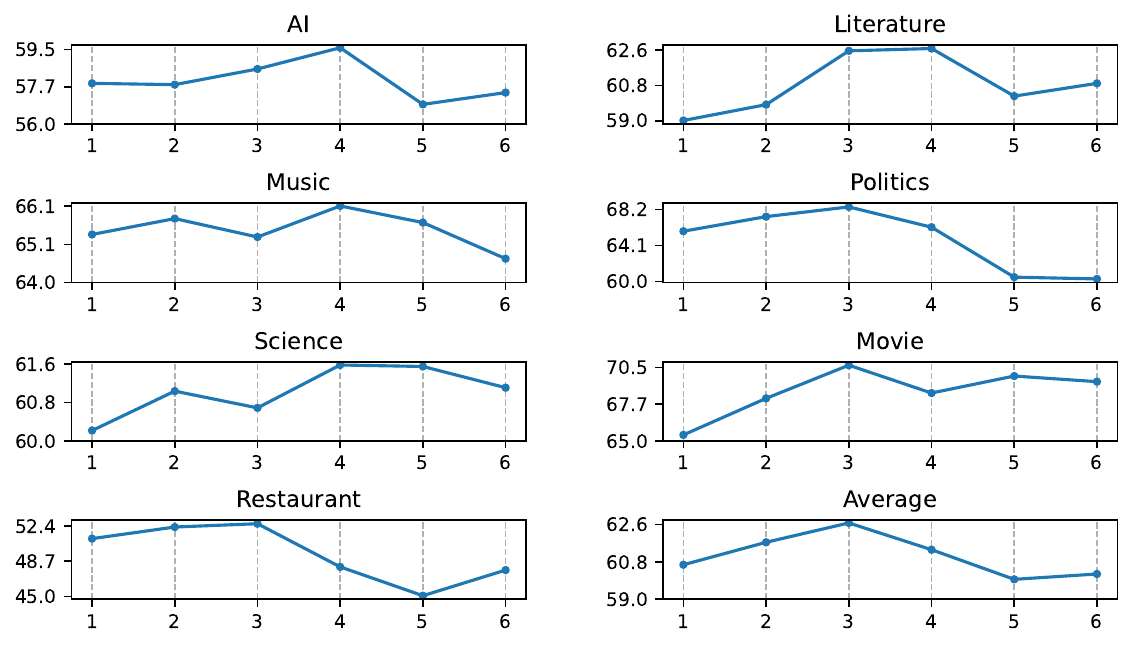}
  \caption{Performance changes with the number of expert models. 
  The horizontal axis is the number of experts for merging, and the vertical denotes the entity-level F1 scores. 
  \textbf{Only using Sampling Evaluation for expert selection.}
  It exhibits similar overall trends to Figure~\ref{fig:expert_num_ds}, but with distinct differences, indicating that both selecting strategies are important and complementary.}
  \label{fig:expert_num_se}
\end{figure*}

Section~\ref{exp:expert_num} explores the relationship between model performance and the number of merged experts. Here, we present results for each strategy, as shown in Figures~\ref{fig:expert_num_ds} and~\ref{fig:expert_num_se}. It can be seen that the two strategies exhibit similar overall trends, but with distinct differences. This further indicates that both strategies are important, highlighting the importance and complementarity of each.

\begin{figure*}[t]
  \centering
  \includegraphics[width=1.0\linewidth]{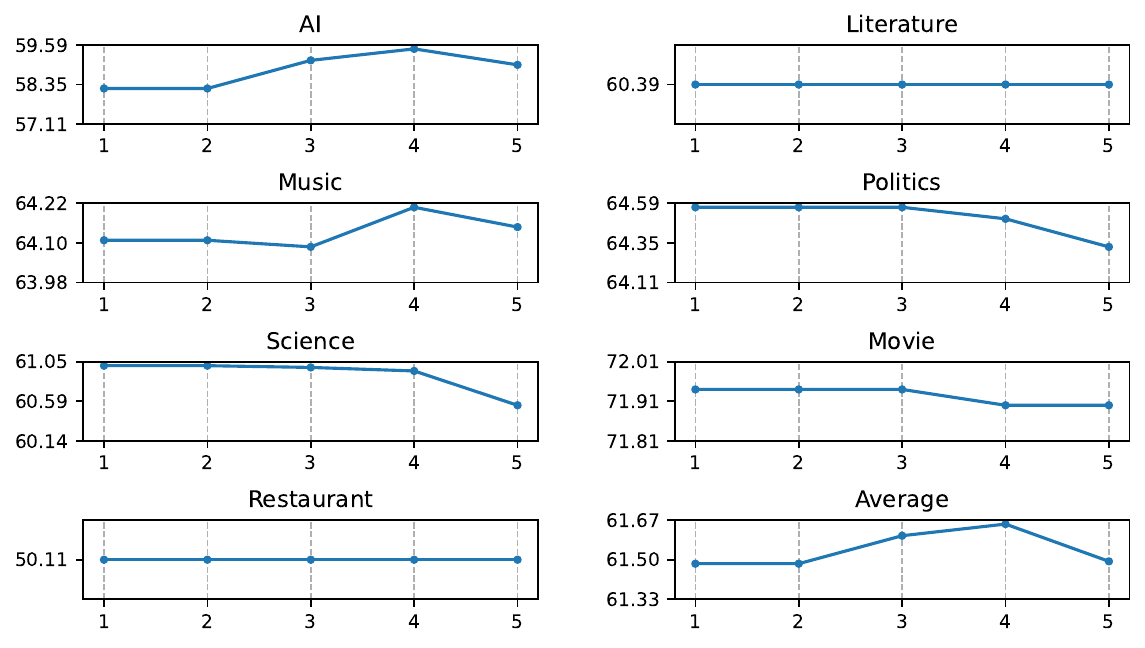}
  \caption{Performance changes with the number of data splits. The horizontal axis is the number of splits, and the vertical denotes the entity-level F1 scores. \textbf{Only using Domain Similarity for expert selection.} It can be seen that the Domain Similarity strategy has a minimal impact, since data from the target domain may be too homogeneous.}
  \label{fig:cluster_ds}
\end{figure*}

\begin{figure*}[t]
  \centering
  \includegraphics[width=1.0\linewidth]{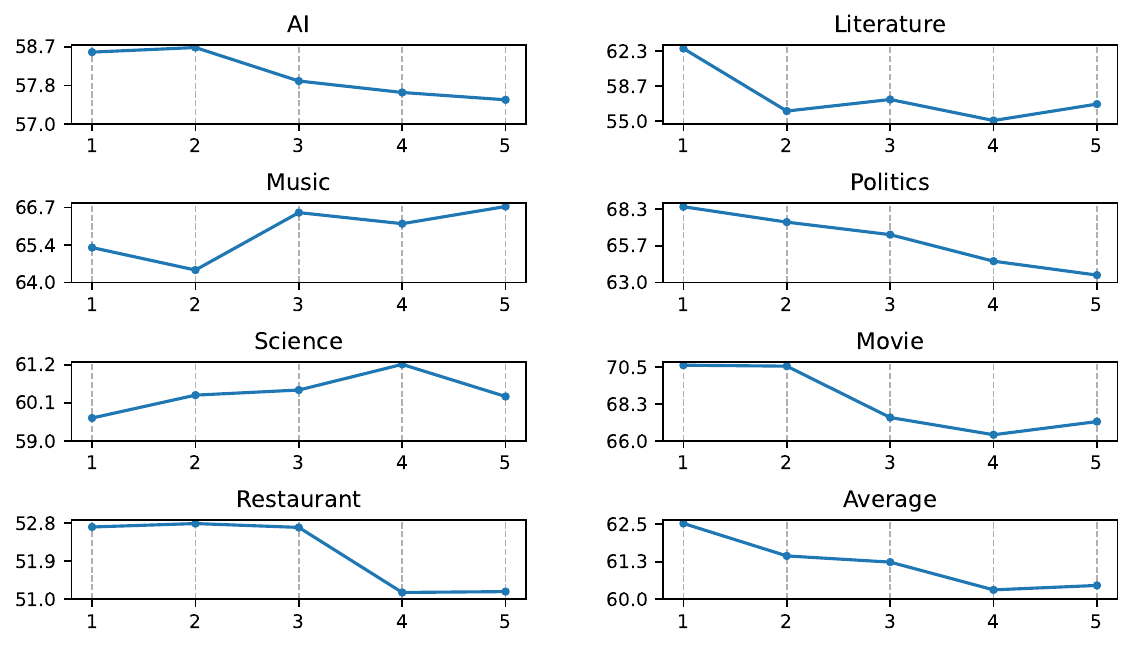}
  \caption{Performance changes with the number of data splits. The horizontal axis is the number of splits, and the vertical denotes the entity-level F1 scores. \textbf{Only using Sampling Evaluation for expert selection.} It shows significant changes when using Sampling Evaluation, with a general downward trend.}
  \label{fig:cluster_se}
\end{figure*}

Section~\ref{exp:finer_adaptation} discusses fine-grained adaptation. Due to the compact nature of the target domain, this does not bring improvements and may even reduce the performance. Here, we analyze the performance of each strategy individually. As shown in Figure~\ref{fig:cluster_ds}, the Domain Similarity strategy has a minimal impact on the subdivision (with a performance difference of less than 0.2 points on average), supporting the hypothesis that the target domain is already too homogeneous to generate distinct splits. In contrast, Figure~\ref{fig:cluster_se} shows significant changes when using Sampling Evaluation, with a general downward trend, which accounts for the overall performance degradation.




\end{document}